\documentclass[a4paper,11pt]{article}

\usepackage[margin = 1.0in]{geometry}

\usepackage{microtype}
\usepackage{url, color, epsfig, amsmath, amsthm, amssymb}

\usepackage{hyperref}

\usepackage{algorithm}
\usepackage[noend]{algorithmic}
\usepackage{amsmath}
\usepackage{enumerate}
\usepackage{thmtools}
\usepackage{thm-restate}
\usepackage{changepage}
\newcommand{\indentone}[1]{
\begin{adjustwidth}{0.2in}{0.1in}
#1 
\end{adjustwidth}
}

\newtheorem{theorem}{Theorem}
\newtheorem*{theorem*}{Theorem}
 
\newtheorem{lemma}[theorem]{Lemma}
 \newtheorem*{definition*}{Definition}

\newtheorem*{example*}{Example}
 \newtheorem*{claim*}{Claim}
\newtheorem*{lemma*}{Lemma}

\renewcommand{\Re}{\mathbb{R}}

\newcommand{\R}{\mathcal{R}}

\newcommand{\B}{\mathcal{B}}

\newcommand{\G}{\mathcal{G}}

\renewcommand{\S}{\mathcal S}

\newcommand{\eps}{\epsilon}
\newcommand{\enet}[1]{$\epsilon$-net}

\def\etal{\emph{et al.}}

\newcommand{\remove}[1]{}

\newcommand{\reel}{\mathbb{R}}

\renewcommand{\S}{\mathcal{S}}

\newcommand{\conv}{\texttt{conv}\ }

\def\db2{\ensuremath{\lfloor d/2 \rfloor}}

\newcommand{\ptas}[1]{\textsc{PTAS}}
\newcommand{\myproof}[1][\textit{Proof}]{\textit{#1. }}

\def\H{\ensuremath{\mathcal{H}}}
\def\R{\ensuremath{\mathcal{R}}}

\def\eps{\ensuremath{\epsilon}}

\def\R{\ensuremath{\mathcal{R}}}

\def\S{\ensuremath{\mathcal{S}}}

\def\H{\ensuremath{\mathcal{H}}}
\def\Re{\ensuremath{\mathbb{R}}}

\def\eps{\ensuremath{\epsilon}}

\DeclareMathOperator{\vcdim}{\mathrm VC-dim}

\usepackage{authblk}

\newtheorem{thmx}{Theorem}
\newtheorem{lemmax}[thmx]{Lemma}

\renewcommand{\Re}{\mathbb{R}}

\title{Optimal Bounds on the VC-dimension\footnote{The research of the first and third authors was supported by the grant ANR SAGA (JCJC-14-CE25-0016-01). The research of the second author was partially supported by the  EPSRC grant no. EP/N019504/1.  }}

\author[1]{M\'onika Csik\'os}
\author[2]{Andrey Kupavskii}
\author[3]{Nabil H. Mustafa}

\affil[1]{\small Karlsruhe Institute of Technology,  Germany. Email: monika.csikos@kit.edu}
\affil[2]{Moscow Institute of Physics and Technology and University of Birmingham. Email:  kupavskii@yandex.ru.}
\affil[3]{Universit\'e Paris-Est,  LIGM,  Equipe A3SI, ESIEE Paris, France. Email: mustafan@esiee.fr.}
\date{}

\begin{document}

\maketitle
\normalsize
\begin{abstract}
The VC-dimension of a set system is a way to capture its complexity and has been a key parameter studied extensively in machine learning and geometry communities. In this   paper, we   resolve two longstanding open problems on bounding the VC-dimension of
two fundamental set systems: $k$-fold unions/intersections of half-spaces, and the simplices set system.
Among other implications, it settles an open question in machine learning that was first studied in the 1989 foundational paper of Blumer, Ehrenfeucht, Haussler and Warmuth~\cite{Bl89} as well
as by Eisenstat and Angluin~\cite{EA07} and Johnson~\cite{J08}.
\end{abstract}

\newpage

\section{Introduction}

Let $(X, \R)$ be a set system, where $X$ is a set of elements
and $\R$ is a set of subsets of $X$. In the theory of learning, the elements of $\R$
are also called \emph{concepts}, and $\R$ is called a \emph{concept class}
on $X$.
For any integer $k \geq 2$, define the \textit{$k$-fold union} of $\R$ as the following set system induced on $X$:
$$\R^{k\cup} = \left\{ R_1 \cup \cdots \cup R_k \colon R_1, \ldots, R_k \in \R \right\}.$$
Similarly, one can define the \emph{$k$-fold intersection} of $\R$, denoted by $\R^{k\cap}$, 
as the set system consisting
of all subsets derived from the common intersection of at most $k$ sets of $\R$.
Note that as the subsets $R_1, \ldots, R_k$ need not necessarily be distinct, we have
 $\R \subseteq \R^{k\cup}$
and $\R \subseteq \R^{k\cap}$.

\paragraph*{Learning theory.} One of the fundamental measures
of `complexity'  of a set system is its \emph{Vapnik-Chervonenkis dimension}, or in short, \emph{VC-dimension}.
 Given a set system $(X, \R)$,
for any set $Y \subseteq X$,
define the \emph{projection} of $\R$ onto $Y$ as 
$$ \R|_{Y} = \left\{ Y \cap R \colon R \in \R \right\}.$$
We say that $\R$ \emph{shatters} $Y$ if
$|\R|_Y| = 2^{|Y|}$; in other words,  any subset of $Y$ can be
derived as the intersection of $Y$ with a set of $\R$.
The \emph{VC-dimension} of $\R$, denoted by $\vcdim(\R)$, 
is the size of the largest subset of $X$ that can be shattered by $\R$.
Originally introduced in statistical learning by Vapnik and Chervonenkis \cite{VC71},
it has turned out to be a key parameter in several
areas, including learning theory, combinatorics and computational geometry.

In  learning theory, the VC-dimension of a concept class measures
the difficulty of learning a concept of the class.
The foundational paper of Blumer, Ehrenfeucht, Haussler and Warmuth~\cite{Bl89}
states that ``the essential condition for distribution-free learnability is finiteness of the Vapnik-Chervonenkis dimension''. 
Among their results, they prove the following theorem.
\begin{thmx}[Blumer \etal~\cite{Bl89}]\label{thm:uniongeneral}
Let $(X, \R)$ be a set system and $k$ be any positive integer. Then
$$ \vcdim \left( \R^{k\cup} \right) = O \Big( \vcdim \left(\R \right) \cdot k \log k \Big),$$
$$ \vcdim \left( \R^{k\cap} \right) = O \Big( \vcdim \left(\R \right) \cdot k \log k \Big).$$
Moreover, there are set systems such that $ \vcdim \left( \R^{k\cup} \right) = \Omega \left( \vcdim \left(\R \right) \cdot k  \right)$ and\\ $ \vcdim \left( \R^{k\cap} \right) = \Omega \left( \vcdim \left(\R \right) \cdot k  \right)$.
\end{thmx}
\noindent
They also considered the question of whether the upper bounds of Theorem~\ref{thm:uniongeneral} are tight in the most basic geometric case when $X \subseteq \Re^d$ is a set of points and $\R$ is the projection of the family of all half-spaces of $\Re^d$ onto $X$. They proved that the VC-dimension of the $k$-fold union of half-spaces in two dimensions is exactly $2k + 1$. For general dimensions $d \geq 3$, they upper-bound the VC-dimension of the $k$-fold union of half-spaces
by $O ( d \cdot k \log k)$. This follows
from Theorem~\ref{thm:uniongeneral} together with the fact that the VC-dimension of the set system
induced by half-spaces in $\Re^d$ is $d+1$. The same upper bound holds for the VC-dimension of the $k$-fold intersection
of half-spaces in $\Re^d$. Later Dobkin and Gunopulos~\cite{DG95} showed that the VC-dimension of
the $k$-fold union of half-spaces in $\reel^3$ is upper-bounded
by $4k$.

Eisenstat and Angluin~\cite{EA07} proved, by giving a probabilistic construction of an abstract set system, that the upper bound of Theorem~\ref{thm:uniongeneral} is asymptotically tight if $\vcdim \left(\R \right) \geq 5$ and that for $\vcdim \left(\R \right)=1$, an upper bound of $k$ holds and that it is tight. A few years later, Eisenstat~\cite{E09} filled the gap by showing that there exists a set system $(X,\R)$ of VC-dimension at most $2$ such that  $\vcdim\left( \R^{k\cup} \right) = \Omega \left( \vcdim \left(\R \right) \cdot k\log k \right)$.

For $d \geq 4$, the current best upper-bound for the $k$-fold union and the $k$-fold
intersection of half-spaces
in $\Re^d$ is still the one given by Theorem~\ref{thm:uniongeneral} almost 30 years ago, while the lower-bound has remained
$\Omega \left( \vcdim(\R) \cdot k \right)$.
We refer the reader to the PhD thesis~\cite{J08} for a summary of the bounds on VC-dimensions
of these basic combinatorial and geometric set systems.  The resolution of the VC-dimension of $k$-fold unions and intersections of half-spaces
is left as one of the main open problems in the thesis. 

\paragraph*{Computational geometry.}
The following set system is fundamental in computational geometry. Given a set $\H$ of hyperplanes
in $\Re^d$, define
$$ \Delta \left(\H\right) = \Big\{ H \subseteq \H \colon \exists \text{ an open $d$-dimensional simplex } \Delta \text{ in $\Re^d$ such that } H = \Delta \cap \H \Big\}.$$
Its importance derives from the fact that it is the set system underlying the construction of \emph{cuttings} via
random sampling  (we refer the reader to Chazelle-Friedman~\cite{CF90}). Cuttings are  
the key tool for fast point-location algorithms and were studied in detail recently by Ezra \etal~\cite{EHKS17}.
They derived the best bounds so far for the VC-dimension of $\Delta(\H)$:

\begin{lemmax}[Ezra \etal ~\cite{EHKS17}] \label{lemma:vcdim_hyperplanes_simplices}
For $d \geq 9$, we have 
$$ d \left(d+1 \right) \leq \vcdim\left( \Delta\left(\H\right) \right) \leq 5 \cdot d^2 \log d.$$
\end{lemmax}

%
%
%

\section{Our Results}

For some time now, it has generally been expected that  $\vcdim \left( \R^{k\cap} \right) = O \left( d  k \right)$
for the $k$-fold unions and intersections of half-spaces.
This upper-bound indeed holds  for a related notion: the \emph{primal shattering dimension} of the $k$-fold unions and intersections of half-spaces is $O(dk)$. 
In fact, as it was pointed out by Bachem~\cite{B18}, several papers in learning theory falsely assume the
same for VC-dimension.
Same for computational geometry literature: for example, the coreset size bounds in the
constructions of~\cite{FL11},~\cite{BEL13}, and \cite{LBK16} would be  incorrect---and require an additional $\log k$ factor in the coreset
size---if the upper-bound
of Theorem~\ref{thm:uniongeneral} was tight for the $k$-fold intersection of half-spaces.
See~\cite{B18} and \cite{BL017} for details.

\medskip 

\noindent In this paper, we completely resolve the question of VC-dimension for the above two set systems.
Our proofs are short and we make an effort to keep them self-contained. 

\medskip
\noindent
\textbf{1.} We show an optimal lower-bound on the VC-dimension
of the $k$-fold union  and the $k$-fold intersection  of half-spaces in $\Re^d$ matching the $O(d \cdot k\log k )$ upper bound of Theorem~\ref{thm:uniongeneral}  
thus settling  affirmatively
one of the main open questions studied
by Eisenstat and Angluin~\cite{EA07}, Johnson~\cite{J08}, and  Eisenstat~\cite{E09}.

\begin{restatable} [Section~\ref{sec:halfspacesunion}]{theorem}{halfspaces}
\label{thm:halfspacesunion}
Let $k$ be a given positive integer and $d \geq 4$ an integer.
Then there exists
a set $P$ of points in $\Re^d$ such that the set system $\R$ induced on $P$ by
half-spaces satisfies
$$ \vcdim \left( \R^{k\cup} \right) = \Omega \Big( \vcdim(\R) \cdot k \log k \Big)
= \Omega \Big( d \cdot k \log k \Big), $$
$$ \vcdim \left( \R^{k\cap} \right) = \Omega \Big( \vcdim(\R) \cdot k \log k \Big)
= \Omega \Big( d \cdot k \log k \Big). $$
\end{restatable}

\indentone{
\noindent \textbf{Remark 1.} This statement also provides a non-probabilistic proof of the lower-bound of Eisenstat and Angluin~\cite{EA07}. }
\vspace{0.1in}

\indentone{
\noindent \textbf{Remark 2.} Observe that if $\overline{\R} :=\{ \reel^d \setminus R :~ R \in \R \}$, then  $\vcdim (\overline{\R}) = \vcdim (\R)$ and 
$$
\vcdim\left(\R^{k\cap}\right) = \vcdim\left(\overline{\R^{k\cap}}\right) = \vcdim\left(\overline{\R}^{k\cup}\right).
$$
  holds by the De Morgan laws. Since for half-spaces $\overline{\R} =\R$, the first
  claim of Theorem \ref{thm:halfspacesunion} implies the second one, i.e., the same lower-bound for  $\R^{k\cap}$, 
  settling another
  question posed by Eisenstat and Angluin~\cite{EA07}. }

\vspace{0.2in}
 
\noindent \textbf{2.} We show an asymptotically optimal bound on the VC-dimension of $\Delta(\H)$,   improving
the bound of Ezra \etal~\cite{EHKS17} and resolving this question that was  studied in the computational geometry community starting in the 1980s.

\begin{restatable}[Section~\ref{sec:cuttings}]{theorem}{cuttings} \label{thm:cuttings}
Let $d \ge 4$ be a given integer. Then there exists a set $\H$ of hyperplanes in $\Re^d$ for which we have
$$\vcdim\left(\Delta\left(\H\right)\right) = \Theta\left( d^2 \log d \right).$$
\end{restatable}

\indentone{
\noindent \textbf{Remark 1.} In fact, we prove a more general result bounding
the VC-dimension of the set system induced by intersection of hyperplanes
with $k$-dimensional simplices in $\Re^d$. See Section~\ref{sec:cuttings} for details.}
 
\bigskip
\bigskip

\noindent \textbf{Organization.} Section~\ref{sec:halfspacesunion} contains the proof of Theorem~\ref{thm:halfspacesunion},
and Section~\ref{sec:cuttings} contains the proof of Theorem~\ref{thm:cuttings}.

\section{Proof of Theorem~\ref{thm:halfspacesunion}.}
\label{sec:halfspacesunion}

We will prove the theorem for $d$ even. The asymptotic lower-bound
for  odd values of $d$ follows from the one  in $\Re^{d-1}$.
For a point $q \in \Re^d$, let $q_i$ denote the $i$-th coordinate value of $q$.

\noindent
The proof will need the following lemma.
\begin{lemma}[\cite{KMP15}]\label{lemma:kmp15}
Let $n, d \ge 2$ be integers.
Then there exists a set $\mathcal B$ of $\lfloor\frac{d}2\rfloor\left(n+3\right)2^{n-2}$ axis-parallel boxes in $\Re^{d}$
such that for any subset ${\mathcal S}\subseteq{\mathcal B}$, one can find a
$2^{n-1}$-element set $Q$ of points in $\Re^d$ with the property that

(i) $Q\cap B\neq \emptyset$ for any $B\in {\mathcal B} \setminus \S$,  and

(ii) $Q\cap B = \emptyset$  for any $B\in \S$.
\end{lemma}
\noindent
Let $d' =  d/2 $.
Apply Lemma~\ref{lemma:kmp15} with $n = \lfloor \log k \rfloor +1$ in $\Re^{d'}$
to get a set $\B$ of boxes in $\Re^{d'}$. 
We assume without loss of generality that the boxes in $\B$ are of the form
$$
B = [x_1, x_1'] \times [x_2,x_2'] \times \dots \times [x_{d'},x_{d'}'], ~\text{ with }~ x_i,x_i' > 0, \ i \in [d'].
$$
For each box $B \in \B$, define the lifted point (see~\cite{PT11})
$$\pi(B) = \left(x_1, \frac{1}{x_1'}, x_2, \frac{1}{x_2'}, \ldots, x_{d'}, \frac{1}{x_{d'}'} \right) \in \reel^{d},$$ 
and let $\pi(\B) := \left\{\pi(B) \colon B \in \B \right\}$. For every $i \in [d]$, let $0 <\alpha_{i,1} < \alpha_{i,2} < \ldots $ denote the sequence of 
distinct values of the $x_i$-coordinates of the elements of $\pi(\B)$. Every such sequence has length at most $|\pi(\B)|$. By re-scaling the coordinates, we can 
assume that 
$$\textrm{for each $i \in [d]$ and $j \leq |\pi\left(\B\right)|$,} \qquad \frac{\alpha_{i,j+1}}{\alpha_{i,j}} > d.$$
Denote the resulting point set by $P$.

We claim that $P$ is shattered by the set system induced by the $k$-fold union of half-spaces in $\Re^d$. To see that, let $P'$ be any subset of $P$. Set $\S$ to be the set of boxes in $\Re^{d'}$ corresponding to $P \setminus P'$.
By Lemma~\ref{lemma:kmp15}, there exists a set $Q$ of
$2^{n-1} = 2^{\lfloor \log k \rfloor} \leq  k$ points in $\Re^{d'}$
such that no box in $\S$ contains any point of $Q$, and each box in $\B \setminus \S$
contains at least one point of $Q$. We will now map each point of $Q \subset \Re^{d'}$ to a half-space in $\Re^d$.

A point $q \in \reel^{d'}$ lies in the box $B = [x_1, x_1'] \times [x_2,x_2'] \times \dots \times [x_{d'},x_{d'}']$ if and only if $x_i \le q_i \le x_i'$ holds for all $i \in [d']$. That happens if and only if the point $\pi(B)$ is contained in the $d$-dimensional box
$$
B(q) = [0, q_1] \times \left[0, \frac{1}{q_1}\right] \times \dots \times [0, q_{d'}] \times \left[0, \frac{1}{q_{d'}}\right].
$$
Let $B(Q) = \left\{B(q) \colon q \in Q\right\}$.
For each box $B \in B(Q)$, we can rescale $B$ if necessary, without changing its intersection with $P$ so that $B$ is of the form
$$
B = [0,b_1] \times [0,b_2] \times \dots \times [0,b_{d}],
$$
where each $b_i$ is equal to $\alpha_{i,j_i}$, for a suitable $j_i$.
Now for each $B \in B(Q)$, we define a half-space $H(B)$ as
\begin{equation}\label{eq:one}
\frac{x_1}{b_1} + \frac{x_2}{b_2} + \dots + \frac{x_{d}}{b_{d}} \le d.
\end{equation}
We claim that for any $p \in P$ and $B \in B(Q)$, $p \in B$ if and only if $ p \in H(B)$. Clearly for any point in $B$, each term on the left-hand side of the inequality \eqref{eq:one} is at most $1$, thus $B \subset H(B)$ and so any point in $B$ lies in $H(B)$. If $p \in P \setminus B$, then $p$ has a coordinate, say $x_i(p)$, that is more than $d$-times larger than $b_i$, which implies $p \not\in H(B)$.
For a point $q \in Q$ let $\rho(q) = H(B(q))$ and $\rho(Q) = \{\rho(q) : q \in Q\}$.
By the properties of the lifting maps $\pi(\cdot)$ and $\rho(\cdot)$:
\begin{align*}
&\text{no box in $\S$ contains any point of $Q$}  \implies \\
& \quad\quad\quad\quad\quad\quad\quad\quad\quad\quad\quad\quad  \text{no point in $ P \setminus P'$ is contained in any half-space of $\rho(Q)$,} \\
&\text{each box in $\B \setminus \S$
contains a point of $Q$}  \implies \\
& \quad\quad\quad\quad\quad\quad\quad\quad\quad\quad\quad\quad \text{each point in $P'$
is contained in some half-space in $\rho(Q)$.}
\end{align*}
In other words, the union of the half-spaces in $\rho(Q)$ contains precisely the set $P'$.
As this is true for any $P' \subseteq P$, the $k$-fold union of half-spaces
in $\Re^d$ shatters $P$. Finally, we have
$$|P| = |\B| = \left\lfloor \frac{d}{2} \right\rfloor ( \lfloor \log k \rfloor +3 ) 2^{\lfloor \log k \rfloor -2}  = \Omega( d \cdot k \log k ),$$
 as desired.
\qed

\section{Proof of Theorem~\ref{thm:cuttings}.}
\label{sec:cuttings}

We prove the following more general theorem from which Theorem~\ref{thm:cuttings} follows immediately by
setting $k = d$.

\bigskip

\noindent Given a set $\H$ of hyperplanes in $\Re^d$, define
$$ \Delta_k \left(\H\right) = \Big\{ H \subseteq \H \colon \exists \text{ an open $k$-dimensional simplex } \Delta \text{ in $\Re^d$ such that } H = \Delta \cap \H \Big\}.$$

\begin{theorem}
For any integer $d\ge 4$ and $k \le d$, there exists a set $\H$ of hyperplanes in $\Re^d$ for which we have
$$\vcdim\left(\Delta_k \left(\H\right)\right) = \Omega \left( d \cdot k \log k \right).$$
\end{theorem} 
\myproof
Apply Theorem~\ref{thm:halfspacesunion} to get  a set $P$ of $\Omega\left(dk\log k\right)$ points in $\reel^d$ such that for every set $P' \subseteq P$, there exists
a set $\G(P')$ of  $k$ half-spaces whose union contains all points in $P'$ and no point in $P \setminus P'$.
From the proof of Theorem~\ref{thm:halfspacesunion} (inequality~(\ref{eq:one})), it follows
that each half-space in $\G (P')$ is of the form:
$$\frac{x_1}{b_1} + \frac{x_2}{b_2} + \dots + \frac{x_{d}}{b_{d}} \le d,$$
where $b_1, \ldots, b_d$ are positive reals. Crucially, this restricted form of half-spaces implies
that each half-space of $\G (P')$ is downward facing, i.e., it contains the origin, which lies below (in the $x_d$-coordinate) its bounding hyperplane.

Using point-line duality~\cite{M02}, map each point $p \in P$ to the hyperplane $H(p)$ by
$$ p = (p_1, \ldots, p_d) \quad \rightarrow \quad H(p):= \{(x_1, \dots, x_d): p_1x_1+p_2x_2+ \dots + p_{d-1}x_{d-1} + p_{d} = x_{d}\}. $$
Let $\H = \left\{H(p) \colon p \in P \right\}$. This is the required set of $\Omega\left(dk\log k\right)$ hyperplanes.
It remains to show that $\H$ is shattered by the set system induced by $k$-dimensional simplices; in other words,
for any $\H' \subseteq \H$, there exists a $k$-dimensional simplex $\Delta$ such that the interior of $\Delta$ intersects each
hyperplane of $\H'$, and no hyperplane of $\H \setminus \H'$.

Fix any $\H' \subseteq \H$ and let $P' = H^{-1} (\H') \subseteq P$ be a set of points in $\Re^d$. Then there
exists a set $\G(P')$ of $k$ half-spaces---each containing the origin---such that the union of the half-spaces in $\G(P')$ contains all
points of $P'$ and no point of $P \setminus P'$. Map each half-space $H \in \G(P')$ to the point $D(H)$ by
$$H :  \frac{x_1}{b_1} + \frac{x_2}{b_2} + \dots + \frac{x_{d}}{b_{d}} \le d \quad \rightarrow \quad D(H) := \left( \frac{b_d}{b_1}, \ldots, \frac{b_d}{b_{d-1}}, d \cdot b_d\right).$$
It is easy to verify that for a point $p\in \Re^d$ and a half-space $H \in \G(P')$, 
$p \in H$ if and only if the point $D(H)$ lies below the hyperplane $H(p)$.
Here we crucially needed the fact
that all half-spaces in $\G(P')$ `face' the same direction, in particular, all are downward facing.

Now consider the $k$ half-spaces in $\G(P')$ and let 
$$D(\G(P')) = \left\{ D(H) \colon H \in \G(P') \right\}$$ 
be $k$ points in $\Re^d$.
As each point $p \in P'$ 
lies in \emph{some} half-space  $H \in \G(P')$, the point $D(H)$ lies below the hyperplane $H(p) \in \H$---or equivalently,
the hyperplane $H(p) \in \H$ has at least one of the $k$ points in the set $D(\G(P'))$ lying below it.
On the other hand, for
each $p \in P \setminus P'$, all the $k$ points in  $D(\G(P'))$ lie
above the hyperplane $H(p) \in \H$.

Finally, consider the convex-hull $\Delta'$ of the $k$ points in $D(\G(P'))$---it is a $(k-1)$-dimensional simplex. 
Now take any hyperplane   $H \in \H$. Then, by the above discussion, $H \in \H'$ if and only if one of these is true: 
\begin{enumerate}
\item $H$ intersects $\Delta'$ and so must have one of its vertices lying below it, or
\item $H$ does not intersect $\Delta'$, and  \emph{all} of its vertices lie below it.
\end{enumerate}

\noindent
Thus   consider the $k$-dimensional simplex
$$ \Delta = \conv \left( D\left(\G\left(P'\right)\right)  \bigcup \left(0, \ldots, 0, \infty\right) \right).$$
Now clearly a hyperplane  $H \in \H$ intersects $\Delta$ if and only if $H \in \H'$.
Note that the point $(0, \ldots, \infty)$ can be any point $(0, 0, \ldots, 0, t)$ for a large-enough value of $t$.
Thus for any $\H' \subseteq \H$, there exists a $k$-dimensional simplex $\Delta$ in $\Re^d$ such
that $\H' = \left\{ H \in \H \colon H \cap \Delta \neq \emptyset \right\}$. This concludes the proof.
 
\qed

%

\pagebreak

\bibliography{kunion}

\end{document}